\documentclass{article}

\usepackage[preprint]{neurips_2026}

\usepackage[utf8]{inputenc} 
\usepackage[T1]{fontenc}    
\usepackage{hyperref}       
\usepackage{url}            
\usepackage{booktabs}       
\usepackage{amsfonts}       
\usepackage{nicefrac}       
\usepackage{microtype}      
\usepackage[usenames,dvipsnames]{xcolor}         
\usepackage{amsmath} 
\usepackage{bbm} 
\usepackage{graphicx}
\usepackage{amssymb}
\usepackage{bm}
\usepackage{subcaption}
\usepackage{caption}
\usepackage{algorithm}
\usepackage{algpseudocode}
\usepackage{float}
\usepackage{wrapfig}

\title{Learning fMRI activations dictionaries across individual geometries via optimal transport}

\author{%
   Sonia Mazelet \\
   CMAP, Ecole Polytechnique \\
   Palaiseau, France\\
   \texttt{sonia.mazelet@polytechnique.edu} \\
   \And
   Rémi Flamary \\
   CMAP, Ecole Polytechnique \\
   Palaiseau, France\\
   \texttt{remi.flamary@polytechnique.edu} \\
   \AND
   Bertrand Thirion \\
   Mind, Inria-Saclay \\
   Palaiseau, France\\
   \texttt{bertrand.thirion@inria.fr} \\
}

\begin{document}

\maketitle

\begin{abstract}
  Dictionary learning is a powerful tool for creating interpretable representations. 
  When applied to functional magnetic resonance imaging (fMRI) data, 
  the resulting patterns of brain activity can be used for various downstream tasks,
  such as brain state classification or population-level analysis. 
  However, a major challenge is the variability in brain geometry across individuals.
  This is usually addressed by projecting each individual brain geometry onto a common template,
  which removes subject-specific information.
  In this work, we introduce a novel approach to dictionary learning on fMRI data that explicitly accounts for this variability.   
  We use the optimal transport-based Fused Gromov-Wasserstein (FGW) distance 
  to compare graphs with different geometries and features. 
  To address the challenge of computing multiple FGW distances for large graphs such as those arising from fMRI data, we
  rely on amortized optimization to learn a
 neural network that predicts an approximation of the optimal transport plans, which substantially reduces the
  computational cost. 
  Additionally, we learn   dictionary atoms that depend on the FGW trade-off parameter, which controls the 
  balance between feature alignment and structural consistency. 
  Numerical experiments on the HCP dataset demonstrate that the proposed approach captures different levels of
  geometric variability in the data and provides representations that preserve essential information.
\end{abstract}

\section{Introduction}

\paragraph{fMRI dictionary learning}
Functional magnetic resonance imaging (fMRI) measures local metabolism throughout the brain 
while subjects perform various cognitive tasks. 
The resulting 4D data are high-dimensional, noisy, and subject-specific, 
which makes learning representations particularly challenging.
Dictionary learning \cite{mairal2009online} provides a natural framework
for this setting, as it represents data as an interpretable combination of atoms and produces compact and denoised representations.
It has been successfully applied to fMRI data 
and outperforms classical decompositions such as Independent component analysis (ICA) \cite{hyvarinen2001independent} 
and Non negative matrix factorization (NMF) \cite{daniel1999learning} on several 
downstream decoding tasks \cite{xie2017decoding,dadi2020fine}. 
Prior work has applied dictionary learning to both task-based \cite{lv2015sparse} and resting-state \cite{eavani2012sparse} 
fMRI under the assumption 
that signals from multiple subjects are registered onto a common geometry. Multi-subject extensions have 
further considered learning shared dictionaries to estimate population-level functional atlases \cite{varoquaux2011multi}.

\paragraph{The problem of inter subject variability} A central challenge in multi-subject fMRI analysis is the substantial 
variability in brain anatomy and functional organization across individuals. 
These differences make it difficult to learn representations that can generalize 
across subjects while being sensitive to individual structure \cite{gordon2017individual,thirion2021from}.
Existing approaches often 
address this issue by projecting all subjects onto a common template geometry,
essentially discarding subject-specific information about the brain structure.  
While practical, such registration procedures may rely on imperfect 
correspondences, distort subject-specific organization, and fail to fully 
preserve inter-subject variability.

\paragraph{Graph dictionary learning} When sampled on surfaces, individual fMRI data naturally admit a graph representation, 
whereby nodes correspond to brain locations and edges encode the neighboring structure between them. 
Dictionary learning was extended to graph-structured data in  \cite{vincent2021online}
through the use of the Optimal Transport based Fused Gromov Wasserstein distance \cite{vayer2020fused},
a powerful tool for comparing graphs with 
different geometries and node features. This framework enables to represent graphs as linear combinations 
of learned graph atoms and has shown promising results on small graph 
datasets, where the learned representations capture both structural and feature 
information.\\ %
Applying graph dictionary learning (GDL) to fMRI data would overcome the 
need to register all subjects on a common geometry, as the FGW distance
explicitly estimates alignments between graphs and therefore allows to learn representations that
account for geometrical variability.
However, the computational cost of the FGW objective remains high, scaling cubically with the number 
of graph nodes. In practice, this restricts existing 
graph dictionary learning approaches to relatively small graphs containing only 
a few tens of nodes \cite{vincent2021online}. As a consequence, direct application of GDL to large-scale 
neuroimaging data such as fMRI graphs, which may involve hundreds or thousands 
of brain locations, remains challenging.


\paragraph{Contributions} We propose Amortized Graph Dictionary learning (AGDL), a new method for FGW graph dictionary learning
that overcomes the two main limitations of FGW-based dictionary learning, which are the high computational cost of FGW 
and the need to choose the trade-off parameter \(\alpha\) between feature and structure alignment.
First, we use amortized optimization \cite{amos2023tutorial} to approximate
the FGW distance, by learning a 
neural network that learns solutions of the FGW problem in an unsupervised way
from pairs of graphs sampled from the dataset. This allows to predict transport plans with a simple forward pass, which substantially reduces the computational cost of GDL and makes it applicable to large graphs.
We use the neural architecture from Unsupervised Learning of Optimal Transport (ULOT) \cite{mazelet2025unsupervised}
to approximate the FGW transport plans.
We also propose to learn dictionaries that explicitly depend on the \(\alpha\) parameter. 
This allows us to capture different levels of geometrical variability in the data and finely select
the parameter with no computational overhead for downstream tasks. 
We apply our method to large scale fMRI data, on the HCP dataset \cite{VanEssen2012hcp}, and show that it learns meaningful 
representations that account for both task-specific and subject-specific information. To the best of our
knowledge, this is the first method for learning joint representations while using individual geometries
on a large scale fMRI dataset.

\section{Fused Gromov Wasserstein distance and its approximation with ULOT}

\subsection{Fused Gromov Wasserstein distance between graphs}

\paragraph{Definition of the FGW distance} A graph is defined as $G=(\bm{F}, \bm{C})$ where $\bm{F} \in 
\mathbb{R}^{n \times d}$ denotes the node feature matrix, $\bm{C} \in \mathbb{R}^{n \times n}$ 
encodes pairwise node distances or relations, typically through an adjacency matrix or a shortest-path matrix.
Given two graphs $G_1$ and $G_2$ an optimal transport plan $\bm{P}(G_1, G_2)$ is a matrix $\bm{P} \in U(n_1, n_2)$ 
for  $U(n_1,n_2)= \{\bm{P} \in \mathbb{R}_+^{n_1 \times n_2} | \bm{P} \mathbbm{1}_{n_2} =\frac{1}{n_1}\mathbbm{1}_{n_1}, \bm{P}^T \mathbbm{1}_{n_1} = \frac{1}{n_2}\mathbbm{1}_{n_2}\}$,
the set of bistochastic matrices, that minimizes the FGW loss $\arg\min_{\bm{P} \in U(n_1, n_2)}  \text{L}^{\alpha} (G_1, G_2, \bm{P})$ \cite{vayer2020fused} for 
\begin{align}
  \text{L}^{\alpha} (G_1, G_2, \bm{P})= &(1-\alpha) \sum_{\substack{i,j=1}}^{n_1,n_2} 
  \left\| \left(\bm{F}_1\right)_i - \left(\bm{F}_2\right)_j \right\|_2^2 P_{i,j} 
  +  \alpha\hspace{-4mm} \sum_{\substack{i,j,k,l=1}}^{n_1,n_2,n_1,n_2}\hspace{-4mm} | 
  \left(\bm{C}_1\right)_{i,k} - \left(\bm{C}_2\right)_{j,l} |^2P_{i,j} P_{k,l}  \label{eq:fugw_loss_gw}.
\end{align} 
The FGW distance is the combination of the Wasserstein term which measures 
discrepancies between node features in the left of \eqref{eq:fugw_loss_gw}, and
the Gromov-Wasserstein term in the right of \eqref{eq:fugw_loss_gw} that captures 
structural distortion between the two graphs. The parameter \(\alpha \in [0,1]\) controls the trade-off between 
feature alignment and structural alignment. 
The FGW distance is a very general distance between graphs that can be 
used to compare graphs with different geometries, features and number of nodes. 
It has been successfully applied to a variety of 
tasks such as graph classification \cite{vayer2020fused}, clustering and dictionary learning \cite{vincent2021online}.

\paragraph{Complexity of solving $\min_{\bm{P} \in U(n_1, n_2)}  \text{L}^{\alpha} (G_1, G_2, \bm{P})$} In order to solve the FGW problem one needs
to minimize the FGW loss with respect to the transport plan $\bm{P} \in U(n_1, n_2)$. Because of the Gromov Wasserstein term, evaluating
the FGW loss for a given plan $\bm{P}$ has a theoretical quartic complexity of \(O(n_1^2 n_2^2)\).
In the case of the square loss, Peyré et al.~\cite{peyre2016gromov} showed that this cost
can be reduced to \(O(n_1 n_2^2 + n_1^2 n_2)\), i.e., cubic complexity. However, this remains 
computationally prohibitive for large graphs, which in our setting typically contain around $1000$ nodes. In the following section, we present a scalable 
solution to this problem based on Unsupervised Learning of Optimal Transport (ULOT) \cite{mazelet2025unsupervised}.

\subsection{Predicting FGW plans with amortized optimization}

\paragraph{Unsupervised learning of optimal transport plans (ULOT)} The FGW objective is non-convex and scales cubically with the number of nodes \cite{peyre2016gromov}, 
which makes direct minimization impractical for large graphs such as those arising in fMRI data. To enable 
efficient computation of transport plans, Mazelet et al. \cite{mazelet2025unsupervised} proposed to learn a neural network conditioned on $\alpha$ for learning approximation \(\bm{P}_\theta^\alpha(G_1,G_2)\)
of the optimal FGW transport plan for graph pairs \((G_1,G_2) \sim \mathcal{D}\), where \(\mathcal{D}\) denotes
a distribution of graphs. They introduced ULOT, a neural network architecture based on cross attention 
and Graph Convolutional Networks. The network is trained using amortized optimization, by directly
minimizing the optimal transport loss, which enables efficient training without
requiring ground truth correspondences. The training loss is
the Fused Unbalanced Gromov-Wasserstein (FUGW) loss (Appendix, eq.~\ref{eq:app_fugw_loss_pen}), a relaxed version of the FGW loss 
that encourages the learned transport plan to satisfy the marginal constraints, with a regularization parameter $\rho$ 
controlling the strength of this encouragement.
Overall, ULOT is trained
by minimizing the expected FUGW loss over graph pairs sampled from the dataset and values of 
$\alpha$ and $\rho$ sampled from a distribution $\mathcal{P}$, which leads to the following optimization problem:
\begin{equation}
  \min_\theta \quad \mathbb{E}_{G_1,G_2\sim\mathcal{D}^2,\alpha,\rho\sim\mathcal{P}} \left[\text{FUGW}^{\alpha, \rho} (G_1,G_2, \bm{P}^{\rho, \alpha}_\theta (G_1,G_2)) \right].
  \label{eq:training}
\end{equation}
At inference, predicting the transport plan between two graphs $G_1$ and $G_2$ is done by a simple 
forward pass through the trained model, which has a complexity of $O(n_1n_2)$, thus enabling the 
application of FGW-based methods to large graphs. 
FUGW has been shown to be pertinent for alignment of fMRI data
\cite{thual2022aligning}, but it also has two parameters that have to be tuned,
potentially individually for each graph pair, which make it impractical for use
a training loss for dictionary learning. Also the marginal relaxation may lead
to loss of information when using it as a loss which is the opposite of what we
want in the context of dictionary learning. So in this work we use ULOT to
predict FGW transport plans with a large value of $\rho$ so that they satisfy the marginal constraints.

\section{Amortized dictionary learning across individual geometries}

\subsection{FGW graph dictionary learning}

\paragraph{FGW graph dictionary learning} The general formulation of FGW dictionary learning introduced in \cite{vincent2021online} is as follows. 
We represent a dataset of graphs $(G_i)_{i \in [N]}$ using a learned dictionary of \textit{atoms} $(G^D_k)_{k \in [K]} = (\bm{F}^D_k, \bm{C}^D_k)_{k \in [K]}$.
Each graph $G_i$ is reconstructed as a linear combination of the dictionary atoms 
$(\sum_{k=1}^{K} \omega^{(i)}_k \bm{F}^D_k, \sum_{k=1}^{K} \omega^{(i)}_k \bm{C}^D_k)$, where $\bm{\omega}^{(i)}=(\omega^{(i)}_k)_{k \in [K]} \in \Sigma_K$ for $\Sigma_K=\left\{ \bm{\omega} \in \mathbb{R}_K^+ | \sum_k  \omega_k = 1 \right\}$ are 
the reconstruction coefficients also referred to as \textit{embeddings}. The FGW dictionary learning problem is defined as
\begin{equation}\label{eq:GDL}
  \min_{\substack{(\bm{\omega}^{(i)})_i \in {\Sigma_K} \\ (\bm{F}_k^D, \bm{C}_k^D)_{k \in [K]}}}
  \frac{1}{N}\sum_{i=1}^{N}
  \min_{\bm{P} \in U(n_i, n)}  \text{L}^{\alpha} (G_i, (\bm{F}(\bm{\omega}^{(i)}), \bm{C}(\bm{\omega}^{(i)})), \bm{P})
\end{equation}
where $\bm{F}(\bm{\omega}^{(i)})=\sum_{k=1}^{K} \omega^{(i)}_k \bm{F}^D_k$ and $\bm{C}(\bm{\omega}^{(i)}) = \sum_{k=1}^{K} \omega^{(i)}_k \bm{C}^D_k$. 
The dictionary is optimized using stochastic gradient updates computed on minibatches of graphs. 
At each stochastic step, the dictionary is first fixed and the unmixing step 
$\min_{\substack{(\bm{\omega}^{(i)})_i \in {\Sigma_K}}}
\frac{1}{N}\sum_{i=1}^{N}
\min_{\bm{P} \in U(n_i, n)}  \text{L}^{\alpha} (G_i, (\bm{F}(\bm{\omega}^{(i)}), \bm{C}(\bm{\omega}^{(i)})), \bm{P})$
is solved to obtain the embeddings for each graph in the minibatch, which are
then used to compute the gradient of the dictionary atoms. Note that another
approach for learning a graph dictionary with a GW barycenter model instead of
linear model has been proposed in \cite{xu2019gromov},
but using a GW barycenter is even more computationally expensive \cite{vincent2021online}.

\paragraph{Complexity of solving GDL} The unmixing step consists in minimizing the FGW 
objective between each graph of the minibatch and its reconstruction, which is 
computed by minimizing the FGW loss with respect to the transport plan $\bm{P} \in U(n_i, n)$ for each graph $G_i$.
Therefore, each unmixing step requires computing the optimal transport plan between each graph of 
the minibatch and its reconstruction $q$ times where $q$ is the number of inner iterations for unmixing. This results in a complexity
$O(pqn^3)$, where $p$ is the number of outer iterations for dictionary updates, $q$ is the 
number of unmixing inner iterations, and $n$ is the number of graph nodes which we
take to be the same for all graphs for simplicity. 
This complexity is prohibitive for large graphs, and therefore limits the application of GDL to 
small graphs with a few tens of nodes, which is several orders of magnitude smaller than the number of nodes in fMRI data, 
which typically contains several thousands of nodes. 
In the next section, we present a scalable solution based on amortized optimization.

\subsection{Amortized dictionary learning on fMRI data}

\paragraph{Common dictionary atom geometry for fMRI data}
In fMRI data analysis, the standard practice is to align individual fMRI data on a common
anatomical template such as the FreeSurfer fsaverage template
\cite{fischl2012freesurfer}. In order to have an interpretable dictionary where
the learned atoms can be visualized and compared across subjects, we also learn
the dictionary atoms on a common geometry represented by the $\bm{C}$ matrix.
While this choice may limit the ability of the model to capture some  subject specificities, 
it still accounts for individual geometrical variability through the FGW optimal transport plan, 
and therefore allows to capture subject specificities in the learned representations.

\paragraph{Amortized dictionary learning} We propose in this paper to optimize an
approximation of the FGW distance at each step of the optimization process
by using amortized optimization \cite{amos2023tutorial} with a pre trained
ULOT model \cite{mazelet2025unsupervised} to predict the transport plan \(\bm{P}_\theta^{\alpha}(G_i,
(\bm{F}(\omega_i), \bm{C}))\)
between each graph and its reconstruction, and then use the neural network model
to predict the OT plan to compute the FGW loss. 
This substantially reduces the computational complexity of GDL to $O(pqn^2)$ and makes it applicable 
to large graphs. The resulting optimization problem is given by:
\begin{equation}\label{eq:unmixing}
  \min_{\substack{(\bm{\omega}^{(i)})_i \in {\Sigma_K}\\ (\bm{F}_k^D)_{k \in [K]}}}
  \frac{1}{N}\sum_{i=1}^{N}
  \text{L}^\alpha \left(
    G_i,(\bm{F}(\bm{\omega}^{(i)}), \bm{C}), \bm{P}_\theta^\alpha(G_i,(\bm{F}(\bm{\omega}^{(i)}), \bm{C}))  \right).
\end{equation}
Since we use a fixed geometry for the atoms, the optimization problem above can be
reformulated as 
\begin{equation}
  \label{eq:GDL_ULOT}
  \min_{\substack{(\bm{\omega}^{(i)})_i \in {\Sigma_K}\\ (\bm{F}_k^D)_{k \in [K]}}} \frac{1}{N} \sum_{i=1}^{N} \sum_{j,k=1}^{n_1,n_2}\left\| \left(\bm{F}_i\right)_j -\bm{F}(\bm{\omega}^{(i)})_k \right\|_2^2 P^\alpha_\theta(G_i, (\bm{F}(\bm{\omega}^{(i)}), \bm{C}))_{j,k}.  
\end{equation}
The problem above is very similar to classical linear DL but uses for each
subject an OT plan predicted by ULOT to align the subject with its
reconstruction. This brings the best of both worlds, that is a simple linear mixture model that 
is easy to optimize and interpret, and the ability to capture subject specificities through 
the use of amortized transport plans.

\paragraph{Learning a dictionary that depends on $\alpha$} In the FGW distance, the $\alpha$ parameter 
controls the trade-off between feature alignment and structural alignment.
Choosing the right value of $\alpha$ is therefore crucial for learning meaningful representations, 
as it determines the extent to which the learned dictionary captures geometrical variability versus feature variability.
However, learning a dictionary from scratch for each value of $\alpha$ would be computationally expensive.
We thus propose to use again amortization and to make the dictionary explicitly depend on $\alpha$,
and obtain representations that capture different  levels of geometrical variability 
with no need to retrain when evaluating the impact of this parameter on downstream tasks.
We first consider a simple model which is a linear interpolation between two endpoint atoms $(\bm{F}^D_0,\bm{F}^D_1)$
with $\alpha$ as interpolation parameter, which allows to capture different levels 
of geometrical variability in the data:
\begin{equation}
  \label{eq:GDL_linear}
  \bm{F}^D = \alpha \bm{F}^D_0 + (1-\alpha) \bm{F}^D_1.
\end{equation}
We also consider a more expressive model with a fully connected MLP that takes $\alpha$ as input and outputs the 
dictionary atoms as follows:
\begin{equation}
  \label{eq:GDL_MLP}
  \bm{F}_\psi^D(\alpha) = \text{MLP}_\psi(\text{softbin}(\alpha)) 
\end{equation}
We encode
each $\alpha$ value in a high dimensional space using soft binning and use the
resulting encoding as input to the MLP. This allows to learn a dictionary that
smoothly evolves with $\alpha$ and therefore can capture different levels of
geometrical variability in the data.


\subsection{Related works}

Traditional dictionary learning \cite{tovsic2011dictionary,mairal2009online} has been successfully applied to 
vectorized fMRI data, where each brain location is treated as a feature and the dictionary is 
learned on the resulting vectorized data \cite{lv2015sparse,shen2017making,iqbal2018shared,khalid2023novel}. 
Large-scale versions using stochastic subsampling have been proposed to learn dictionaries 
on large fMRI datasets \cite{mensch2018stochastic,dadi2020fine}.
Extensions have been introduced on graph structured data where the same structure is shared among all
samples and the dictionary is learned on this shared structure
\cite{thanou2014learning}.
However, these methods cannot account for the variability in brain geometry
across subjects, which is a crucial aspect of fMRI data.
In this work, we train on the individual geometry of the subject thanks to the
use of OT that provides an alignment between structures. This is a crucial aspect of 
fMRI data as it encodes the spatial organization of brain activity.

Our method builds on top of Graph Dictionary learning from \cite{vincent2021online}, which
introduced dictionary learning for graphs using the FGW distance. However this method
relied on exact solvers for the FGW distance which makes the method inapplicable to large graphs. 
Our method overcomes this limitation by using amortized optimization to learn an approximation of the FGW transport plans, which substantially reduces the computational cost and makes it applicable to large graphs. \\
Another important contribution is that our method learns a dictionary that explicitly 
depends on the $\alpha$ parameter, which allows to easily compute
dictionary atoms for different values of the trade off parameter. This is
particularly interesting because the method in \cite{vincent2021online} learns a single
dictionary for a fixed value of $\alpha$, so exploring multiple values of $\alpha$ would require many training runs.\\
Moreover, we demonstrate the application of our method to fMRI data, 
which is a particularly challenging setting due to the large size of the graphs and the high variability across subjects. 
In contrast, the method in \cite{vincent2021online} was
only evaluated on small synthetic or molecule graph datasets.

\section{Numerical experiments on fMRI data}

\subsection{Experimental setting}


\paragraph{HCP dataset and DL baseline} We train ULOT and perform dictionary learning on the HCP dataset \cite{van2013wu} which consists in 
fMRI data of 1200 subjects performing a variety of tasks. We use the task fMRI data for 
7 different tasks (emotion, gambling, motor, relational, language, social and working memory).
The dataset contains two types of brain geometries for each subject, the common \textit{fsaverage} geometry shared across individuals, and the 
individual geometry that is specific to each subject in terms of brain shape and folding. 
For each of these geometries, we build the dataset on the left hemisphere by running a General Linear Model (GLM) using the Nilearn 
library \cite{Nilearn} on the 
task fMRI data to obtain maps for the main contrasts in each subject. A contrast compares brain activity between 
two task conditions, highlighting the brain areas that are more or less active for the task of interest.
For each subject we compute a maximum of $10$ different contrast maps,
depending on which task signal is available. 
We detail the contrast computations in Appendix \ref{sec:contrasts}.
\newline Each contrast map is then represented as a graph, 
where nodes correspond to positions on the brain surface and edges encode the geodesic distance 
between these positions. Graphs are built by aggregating positions on the brain surface using the 
Ward algorithm \cite{thirion2014fmri}. The node features are the average
contrast value in each cluster concatenated with the 3d position of the node. Therefore,
since we have approximately $10$ contrasts per subject, we get an approximate total of $12000$ graphs.
We split the subjects into $70/15/15 \%$ train/validation/test sets, and we use the same split for both 
the ULOT training and the dictionary learning.\\
As a baseline, we perform dictionary learning directly on the fMRI data projected 
onto the \textit{fsaverage} common geometry, which is the standard practice in fMRI data analysis.
In this setting, the reconstruction loss can be evaluated 
without solving an additional graph-alignment problem, and reduces to a standard
least-squares objective. The proposed AGDL is done
 by training on native geometry graphs, and the atoms are defined on the common
 geometry.

\begin{figure}[t]
  \centering
  \includegraphics[width=\textwidth]{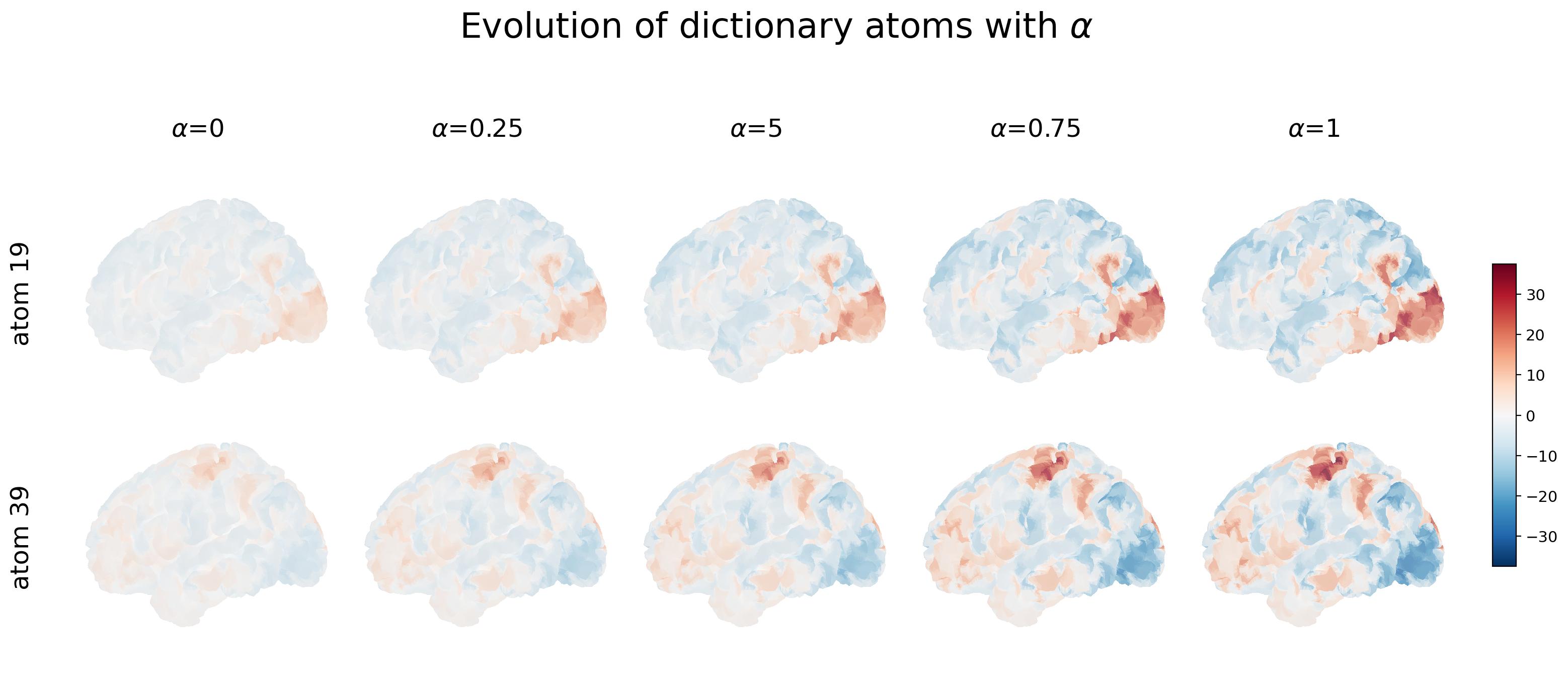}
  \caption{Visualization of the learned dictionary atoms for different values of $\alpha$ from
  $0$ (pure Wasserstein) to $1$ (pure Gromov-Wasserstein). The top row atom is the atom most correlated with 
  contrast \textit{fear - neutral} and the bottom row atom is the most correlated with contrast \textit{right hand - average}.
  }
  \label{fig:atoms}
\end{figure}
\paragraph{Pre training for amortized optimization of the OT plans} For amortized optimization of 
the FGW transport plans, we 
first pre-train a ULOT model on the HCP dataset by minimizing the FUGW loss \eqref{eq:training} on pairs of 
graphs using either native-common or native-native pairs. We train the model stochastically by sampling
$200\,000$ graph pairs per epoch
and training for $105$ epochs (see Appendix \ref{sec:ULOT_convergence}). We fix a batch size of $64$ and a learning rate of $0.002$. 
We sample $\alpha$ values from a Beta distribution with parameters $(0.5,0.5)$ and $\rho$ values from a log-uniform distribution 
on $[10^{-7}, 1]$. The model is then frozen for use in the subsequent
dictionary learning step. For more
details on the experimental setting and convergence analysis for
ULOT pre training, see Appendix \ref{sec:experimental_details}.

\begin{wrapfigure}{r}{0.3\textwidth}
  \centering \vspace{-0.3cm}
  \includegraphics[width=0.2\textwidth]{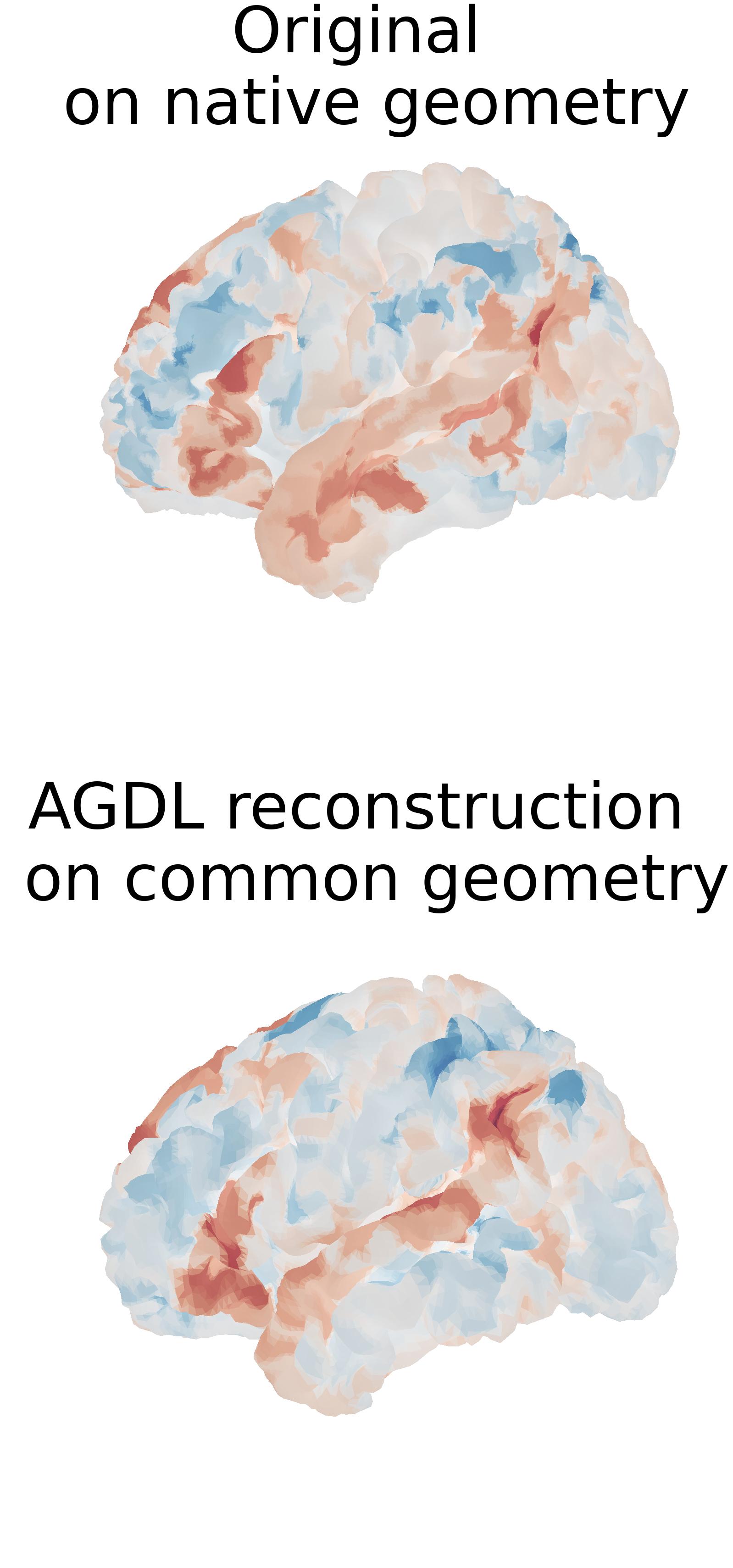}
\caption{Original contrast (top) and its reconstruction (bottom) for a given subject and contrast \textit{story - math}.
}
\centering
\label{fig:reconstruction} \vspace{-1.1cm}
\end{wrapfigure}
\paragraph{AGDL} We train AGDL with both the linear interpolation \eqref{eq:GDL_linear} and the MLP \eqref{eq:GDL_MLP} on the HCP dataset by minimizing the loss in 
Eq.~\eqref{eq:GDL_ULOT}, using a batch size of $32$, a learning rate of $0.001$, and training for $1000$ epochs.
We train the model stochastically by sampling $100$ graphs per epoch.
The atoms are initialized with graphs from the HCP dataset defined on the common \textit{fsaverage} geometry, 
and we use graphs of size \(950\) for both the dictionary atoms and the training, validation and testing datasets. 
We consider a dictionary with \(K=40\) atoms, which is $4$ times the number of contrasts, in order to get a model 
expressive enough to represent different levels of (subject) variability for each contrast. 
The MLP model for the dictionary conditioned by $\alpha$
has two layers
with hidden dimension $16$. For training, the values of \(\alpha\) are sampled uniformly from \([0,1]\). 
For more details on the experimental setting for
dictionary learning see Appendix \ref{sec:experimental_details}.

\subsection{Visualisation of the learned dictionary and reconstructions}


\paragraph{Reconstruction} Figure \ref{fig:reconstruction} shows the 
original contrast \textit{story - math} for a given subject and its 
reconstruction for a given subject and contrast. This figure highlights
the difference between the native and common geometry, both in terms 
of brain shapes and foldings.
We see that the reconstruction 
captures the main activation patterns of the original contrast, 
while being smoother and less noisy.

\paragraph{Evolution of the dictionary atoms with $\alpha$} We learn dictionary atoms that depend
on the FGW parameter $\alpha$. At inference time, this allows to very quickly obtain dictionaries 
that capture different tradeoffs between a reconstruction that focuses on feature alignment ($\alpha=0$) 
and a reconstruction that focuses on structural alignment ($\alpha=1$). Figure \ref{fig:atoms} shows 
the learned dictionary atoms for different values of $\alpha$ from $0$ (pure Wasserstein) to
 $1$ (pure Gromov-Wasserstein). We choose the atoms the most correlated with contrast
\textit{fear - neutral} (first row) and \textit{right hand - average} (second
row) for visualization.\\
We see that the atoms have localized activations that correspond to the contrasts chosen.
Moreover, increasing $\alpha$ leads to sharper atoms.
For a visualization of all AGDL learned atoms see Appendix \ref{sec:atoms} Figure \ref{fig:al_atoms}.

 \begin{wrapfigure}{r}{0.5\textwidth}\vspace{-0.5cm}
  \centering
  \includegraphics[width=0.5\textwidth]{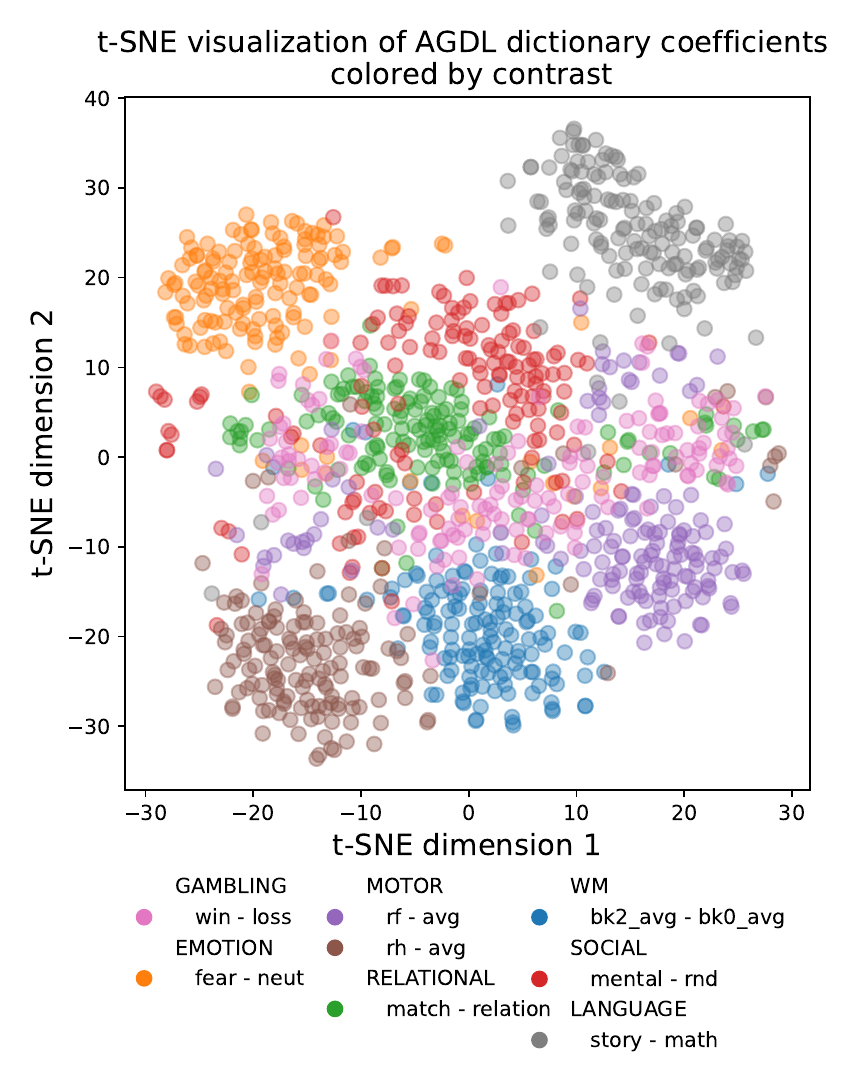}
\caption{t-SNE visualization of the learned embeddings for $\alpha=0.22$ (best contrast classification performance). 
Each point corresponds to a graph, colored according to its contrast.}
\label{fig:tsne}\vspace{-1.5cm}
\end{wrapfigure}
\begin{figure}[t]
  \centering
  \includegraphics[width=0.8\textwidth]{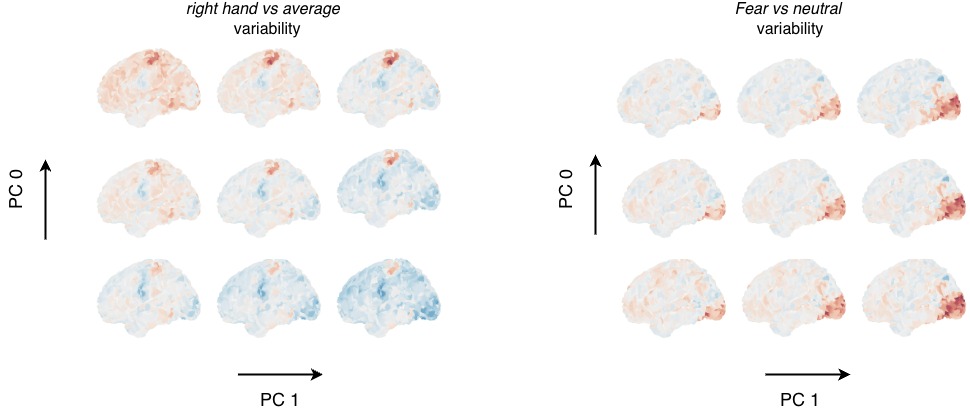}
  \caption{Visualization of the reconstructions obtained by moving along the first two principal 
  components around the average in the embedding space for contrast \textit{right hand - average} (left) and \textit{fear - neutral} (right).}
\label{fig:pc_grid}
\end{figure}

\paragraph{PCA in the embedding space} To further illustrate subject variability in the learned representation, 
we choose two classical contrasts \textit{right hand - average} and \textit{fear - neutral} and perform PCA in the embedding 
space across many subjects. 
We then visualize in Figure \ref{fig:pc_grid} how the reconstruction 
evolves when moving around the mean along the principal components in 
the embedding space. This illustrates the subject-specific variability 
for fixed contrasts.

\subsection{Dictionary embeddings interpretation}

The fMRI graphs are characterized by their subject identity and their contrast. It is therefore important 
to build representations that capture both subject-specific and contrast-specific information. 
In the following we evaluate the ability of our method to capture both types of information.

\begin{figure}[t]
  \centering
  \includegraphics[width=\textwidth]{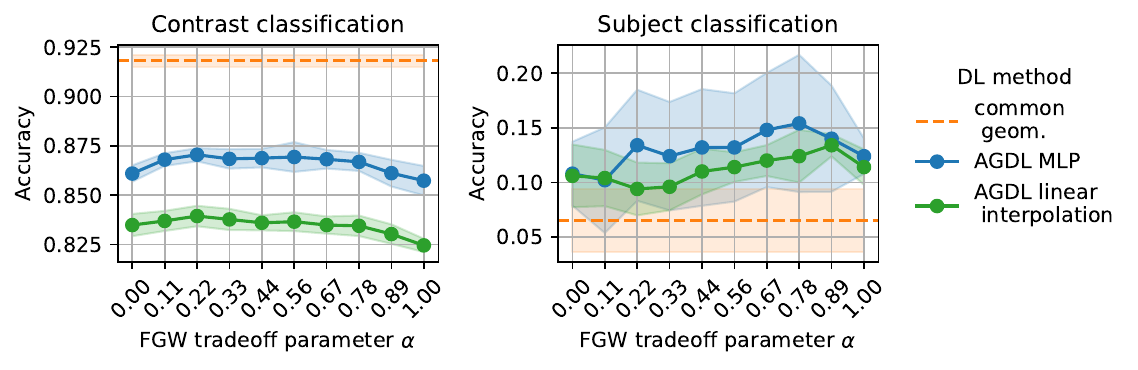}
  \caption{Performance for contrast classification (left) and subject classification (right) 
  for different values of the FGW tradeoff parameter $\alpha$ from $\alpha=0$ (Wasserstein)
  to $\alpha=1$ (Gromov). We compare the baseline DL on common geometries, with AGDL with 
  linear interpolation and AGDL with MLP.}
  \label{fig:classification}
\end{figure}

\paragraph{Discriminative power of embeddings on contrasts} We evaluate 
the ability of the learned embeddings to discriminate between different task contrasts, which 
is a common downstream task in fMRI data analysis.
Figure \ref{fig:tsne}, shows a t-SNE visualization of the 
embeddings obtained with $\alpha=0.22$, where each point represents a graph embedding colored 
according to its contrast. We observe that the learned representations capture meaningful information about the 
contrasts, as graphs corresponding to the same contrast tend to cluster together in the embedding space.
Since we learn embeddings for the left hemisphere, 
we do not consider here contrasts \textit{left hand - average} and \textit{left foot- average}, as
they mostly activate the right hemisphere. For a visualization of the embeddings for all contrasts see
Appendix \ref{sec:embeddings} Figure \ref{fig:tsne_all}.
\\We also evaluate the performance of the learned representations on a contrast
classification task, for different values of $\alpha$, where 
we train an SVM classifier with Gaussian kernel to predict the task contrast from the learned embeddings. We show in Figure 
\ref{fig:classification} (left) the performance of the classifier for different values of $\alpha$, and 
observe that the performance depends on $\alpha$ with the best performance obtained for  $\alpha=0.22$, which corresponds
 to a balance between Gromov and Wasserstein distance. This underlines the fact that capturing both feature and structural 
 information is important for discriminating between contrasts.
However, AGDL does not perform as well as the 
baseline for contrast classification. We believe that this comes from the fact
that the  embeddings also encode individual geometrical information which is
noise for contrast classification, whereas the baseline cancels out this
variability by projecting all subjects onto a common geometry before DL.
However, the MLP model clearly performs better than the linear interpolation model, 
suggesting that greater expressivity is needed to learn good representations.

\begin{figure}[t]
  \centering
  \includegraphics[width=0.8\textwidth]{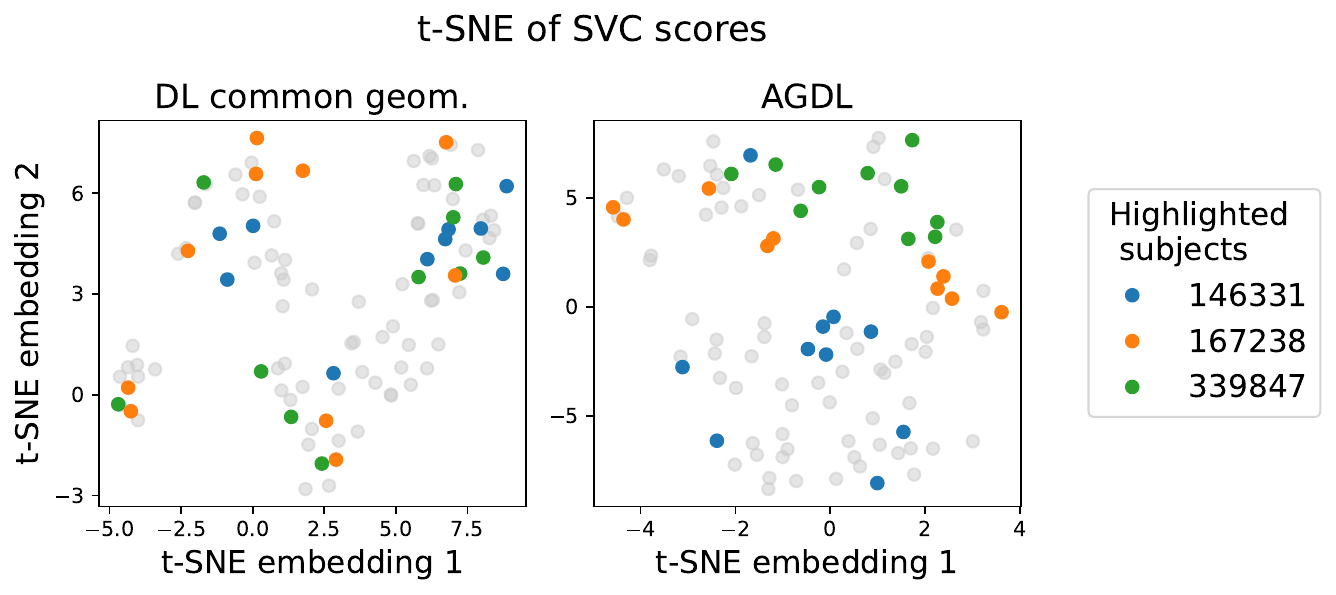}
\caption{t-SNE visualization of the training set SVC scores for $\alpha=0.78$ (best performance for subject classification),
trained on $10$ subjects. $3$ random subjects are highlighted to underline the clustering of the SVC scores.}
\label{fig:tsne_subjects}
\end{figure}

\paragraph{Discriminative power of embeddings on subjects} 
We further evaluate  whether the learned embeddings can discriminate between subjects, 
which is a more challenging task that requires  the embeddings to capture subject-specific information.
This information is typically partially lost when projecting all subjects onto a common geometry. 
We evaluate the performance of the learned representations on a subject classification 
task, where we train a support vector classifier (SVC) to predict the subject from the learned embeddings. 
For each value of $\alpha$, we randomly select 10 subjects, train the SVC on the embeddings of the graphs 
associated with those subjects, and evaluate it on the same subjects under different contrasts. 
This procedure is repeated over 5 random seeds for each value of $\alpha$, and we report the average performance.
We show in Figure \ref{fig:classification} (right) the performance of the classifier for different values 
of $\alpha$, and observe that the best performance is obtained for $\alpha=0.78$, which corresponds to a setting where the 
learned representations capture both feature and structural information, but
interestingly different to the optimal $\alpha$ for contrast classification. These results highlight the strength of our method
that accounts for subject-specific information by performing dictionary learning directly in the native geometries.
\newline We visualise the subject-specific information captured in the learned representations in Figure \ref{fig:tsne_subjects} by performing a 
t-SNE visualization on the DL and AGDL SVC scores, with $\alpha=0.78$. 
We observe that the learned representations captures subject-specific 
information, as the SVC scores corresponding to the same subject tend to cluster together in the embedding space.
In comparison, the SVC for the baseline does not clearly cluster graphs from the same subject together. This result is particularly interesting
because it shows that the individual geometrical information is essential for learning good representations.

\begin{figure}[t]
  \centering
  \includegraphics[width=\textwidth]{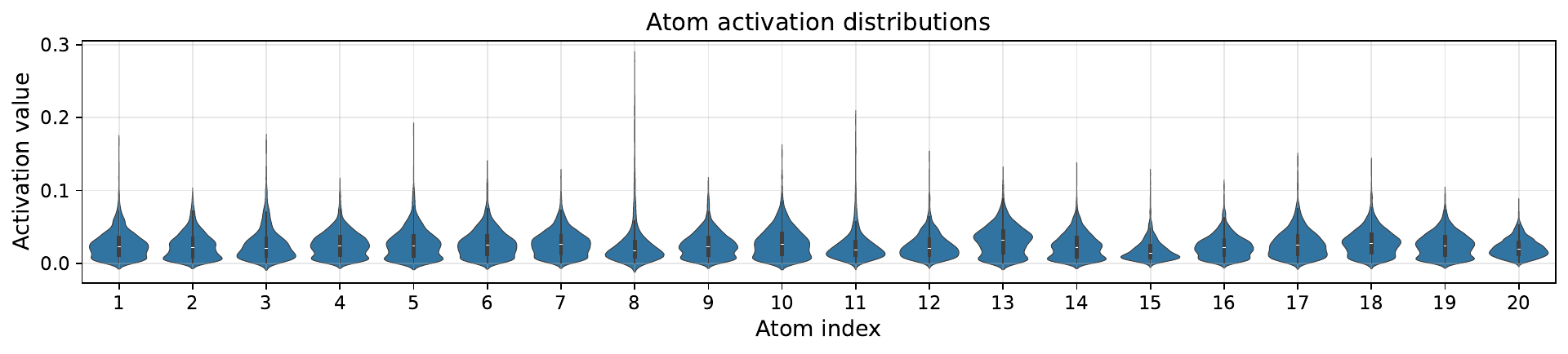}
  \vspace{-5mm}
  \caption{Distribution of the activations for the first $20$ atoms.}
  \label{fig:activations}
\end{figure}

\paragraph{Dictionary activations visualization} Figure \ref{fig:activations} shows the distribution 
of the activations. The first $10$ atoms are the atoms most correlated with each of the $10$ contrasts and
the last $10$ atoms are randomly sampled from the remaining ones. We see that all atoms are
activated for a large number of graphs, which suggests that the model is using all atoms 
to reconstruct the data in average. Moreover, the activations are not binary, which suggests that the complexity
of the data requires the use of multiple atoms for individual contrasts reconstruction.


\section{Conclusion}

We propose a new method, AGDL, for doing optimal transport based graph dictionary
learning on large fMRI graphs using 
amortized optimization with a pre-trained ULOT model to predict the transport 
plans used in the FGW loss and learning dictionaries conditioned on the 
optimal transport loss parameter $\alpha$.
We show that this method  can be applied to large fMRI data and learns meaningful representations
that capture both  task-specific and subject-specific information. {We implement
two different dictionary models, one based on a simple linear 
interpolation, and one based on a more expressive 
MLP. We find that the MLP performs better on downstream classification tasks, 
suggesting that greater expressivity is needed to learn useful representations.} 
\\This method goes beyond the common practice of  projecting all subjects onto a common geometry, 
which biases the tight coupling between function and cortical geometry and thus degrades subject-specific information.
AGDL is in the line of \cite{thual2022aligning} that showed that 
projecting to a common geometry degrades subject-specific information, and that 
using FGW-based methods can help to capture this information. We go a step further by proposing a 
method for learning representations directly from the native geometries. To our knowledge, this is the 
first work to propose a scalable joint representation of graphs of different geometries while
keeping information about their individual geometries. 
\newline This work is a first attempt to perform DL from fMRI data with
heterogeneous geometries. As such, it has limits and opens the door to several potential extensions.
  First, we will in future work learn a joint dictionary on the right and left
  hemisphere which will improve the performance in contrast classification, especially for the contrasts
that mostly activate the right hemisphere. {Second, we will explore the impact of relaxing 
the fixed geometry assumption for atoms which should produce more expressive atoms 
and therefore better embeddings. This would imply losing the surface geometry of the atoms 
  which is a key aspect of interpretability in fMRI data, but it would be
  interesting to see how to interpret the learned geometries for the atoms.}
Finally, we could learn the atoms and the reconstruction coefficients in a 
fully amortized way following the method in \cite{amos2023tutorial} by learning a neural network that takes as input a graph and outputs its embedding, which would further reduce the computational cost of the method and make 
it applicable to even larger datasets and for near real time processing of fMRI data. 

\begin{ack}
  This work was granted access to the HPC resources of IDRIS under the allocation 2025-AD011016350R1 made by GENCI. 
  This work is supported by Hi! PARIS and ANR/France 2030 program (ANR-23-IACL-0005).
  This research was also supported in part by the French National Research Agency (ANR) through the  MacLeOD project (ANR-25-PEIA-0005).
  This work benefited from state aid managed by the Agence Nationale de la Recherche under the France 2030 programme, 
  reference ANR-22-PESN-0012 and from the European Union’s Horizon 2020 Framework Programme for Research and Innovation under the Specific 
  Grant Agreement HORIZON-INFRA-2022-SERV-B-01, grant agreement number 10.3030/101147319.
  Finally, it received funding from the Fondation de l'École polytechnique.
\end{ack}

\medskip

\newpage

\bibliographystyle{plain}
\bibliography{biblio}


\appendix

\section{AGDL algorithm and amortized ULOT training}
\subsection{AGDL algorithm}

The algorithm used for AGDL is described in Algorithm \ref{alg:agdl}, and the unmixing 
step is described in Algorithm \ref{alg:agdl-unmixing}. In practice, during unmixing 
we parameterize the simplex coefficients as $\omega^{(i)} = \text{Softmax}(z^{(i)})$ and 
optimize the logits $z^{(i)}$ by gradient descent. At inference time for dictionary learning we fix $\rho=0.9$.

\begin{algorithm}[t]
  \caption{AGDL Dictionary Learning with ULOT-based Unmixing}
  \label{alg:agdl}
  \begin{algorithmic}[1]
  
  \Require Dataset $(G_i)_{i \in [N]}$ with $G_i=(\bm{F}_i,\bm{C}_i)$, initial dictionary atoms $(\bm{F}_k^{D,(0)})_{k \in [K]}$, common geometry $\bm{C}$, FGW parameter $\alpha$, batch size $B$, unmixing tolerance $\varepsilon$, unmixing step size $\eta_{\omega}$, dictionary step size $\eta_D$, unmixing iterations $T$, outer iterations $S$
  
  \For{$s = 0, \dots, S-1$}
      \State Sample minibatch indices $\mathcal{I}^{(s)} \subset [N]$ with $|\mathcal{I}^{(s)}| = B$
      \State $(\bm{\omega}^{(i,s)})_{i \in \mathcal{I}^{(s)}} \gets \textsc{Unmixing}\!\left((\bm{F}_k^{D,(s)})_{k \in [K]}, \bm{C}, (G_i)_{i \in \mathcal{I}^{(s)}}, \alpha, \varepsilon, \eta_{\omega}, T\right)$
      \For{$i \in \mathcal{I}^{(s)}$}
      \Statex
      \[
      \bm{F}^{(s)}\!\left(\bm{\tilde{\omega}}^{(i,s)}\right) \gets \sum_{k=1}^K \tilde{\omega}_k^{(i,s)} \bm{F}_k^{D,(s)}
      \]
      \Statex
      \[
      \bm{P}_i^{(s)} \gets \bm{P}_\theta^\alpha\!\left(G_i,\left(\bm{F}^{(s)}\!\left(\bm{\tilde{\omega}}^{(i,s)}\right), \bm{C}\right)\right)
      \]
      \EndFor
      \State \textbf{Batch dictionary loss:}
      \Statex
      \[
      \mathcal{J}_{\mathcal{I}^{(s)}} \gets \sum_{i \in \mathcal{I}^{(s)}}
      \left[
      \text{L}^\alpha\!\left(
      G_i,\left(\bm{F}^{(s)}\!\left(\bm{\tilde{\omega}}^{(i,s)}\right), \bm{C}\right), \bm{P}_i^{(s)}
      \right) 
      \right]
      \]
      \Statex
      \[
      \left(\bm{F}_k^{D,(s+1)}\right)_{k \in [K]}
      \gets
      \left(\bm{F}_k^{D,(s)}\right)_{k \in [K]}
      -
      \eta_D \nabla_{(\bm{F}_k^D)_{k \in [K]}} \mathcal{J}_{\mathcal{I}^{(s)}}
      \]
  \EndFor
  
  \State \Return $(\bm{F}_k^{D,(S)})_{k \in [K]}$
  
  \end{algorithmic}
  \end{algorithm}

\begin{algorithm}[t]
  \caption{UNMIXING}
  \label{alg:agdl-unmixing}
  \begin{algorithmic}[1]
  
  \Require Dictionary atoms $(\bm{F}_k^D)_{k \in [K]}$, common geometry $\bm{C}$, minibatch $(G_i)_{i \in \mathcal{I}}$ with $G_i=(\bm{F}_i,\bm{C}_i)$, FGW parameter $\alpha$, tolerance $\varepsilon$, step size $\eta$, max iterations $T$
  
  \State Initialize $\bm{z}^{(i,0)} \gets \mathbbm{0}_K$ for all $i \in \mathcal{I}$
  \State $t \gets 0$, $\Delta \gets +\infty$
  \While{$\Delta > \varepsilon$ \textbf{and} $t < T$}
      \For{$i \in \mathcal{I}$}
      \Statex
      \[
      \bm{\omega}^{(i,t)} \gets \text{Softmax}\!\left(\bm{z}^{(i,t)}\right)
      \]
  
      \State \textbf{Reconstruction:}
      \Statex
      \[
      \bm{F}\!\left(\bm{\omega}^{(i,t)}\right) \gets \sum_{k=1}^K \omega_k^{(i,t)} \bm{F}_k^D
      \]
      \State \textbf{Predicted transport plan:}
      \Statex
      \[
      \bm{P}_i^{(t)} \gets \bm{P}_\theta^\alpha\!\left(G_i,\left(\bm{F}\!\left(\bm{\omega}^{(i,t)}\right), \bm{C}\right)\right)
      \]
      \EndFor
      \State \textbf{Batch loss:}
      \Statex
      \[
      \mathcal{L}_{\mathcal{I}}^{(t)} \gets \sum_{i \in \mathcal{I}}
      \left[
      \text{L}^\alpha\!\left(
      G_i,\left(\bm{F}\!\left(\bm{\omega}^{(i,t)}\right), \bm{C}\right), \bm{P}_i^{(t)}
      \right) 
      \right]
      \]
      \State \textbf{Gradient step:}
      \Statex
      \[
      \left(\bm{z}^{(i,t+1)}\right)_{i \in \mathcal{I}}
      \gets
      \left(\bm{z}^{(i,t)}\right)_{i \in \mathcal{I}}
      -
      \eta \nabla_{(\bm{z}^{(i)})_{i \in \mathcal{I}}} \mathcal{L}_{\mathcal{I}}^{(t)}
      \]

      \Statex
      \[
      \Delta \gets \left(\sum_{i \in \mathcal{I}} \left\|\bm{z}^{(i,t+1)} - \bm{z}^{(i,t)}\right\|_2^2 \right)^{1/2}
      \]
      \State $t \gets t + 1$
  
  \EndWhile
  
  \State \Return $(\text{Softmax}(\bm{z}^{(i,t)}))_{i \in \mathcal{I}}$
  
  \end{algorithmic}
  \end{algorithm}

  \subsection{Fused unbalanced Gromov-Wasserstein loss for amortized optimization of the transport plans}

  The ULOT model is trained by minimizing the Fused Unbalanced Gromov-Wasserstein (FUGW) loss, 
  which is a generalization of the Fused Gromov-Wasserstein (FGW) distance that allows to
  compare graphs with different total mass and to learn transport plans that are not necessarily balanced.
  In this setting, graphs are defined with additional node weights $\omega$ representing their mass.
  We fix uniform weights ${\omega}=1/n$. The Fused Unbalanced Gromov-Wasserstein (FUGW) loss is defined as follows:
  \begin{align}
    \text{FUGW}^{\alpha,\rho} (G_1, G_2, \bm{P})= &(1-\alpha) \sum_{\substack{i,j=1}}^{n_1,n_2} \left\| \left(\bm{F}_1\right)_i - \left(\bm{F}_2\right)_j \right\|_2^2 P_{i,j} \label{eq:app_fugw_loss_w} \\ &+  \alpha \sum_{\substack{i,j,k,l=1}}^{n_1,n_2,n_1,n_2} | \left(\bm{C}_1\right)_{i,k} - \left(\bm{C}_2\right)_{j,l} |^2 P_{i,j} P_{k,l} \label{eq:app_fugw_loss_gw}
    \\  &+ \rho \left( \text{KL}(\bm{P}_{\#1}  \otimes \bm{P}_{\#1} | \bm{\omega}_1 \otimes  \bm{\omega}_1)  + \text{KL}(\bm{P}_{\#2}  \otimes \bm{P}_{\#2} | \bm{\omega}_2 \otimes  \bm{\omega}_2) \right), \label{eq:app_fugw_loss_pen}
  \end{align} 
  where $\bm{P}_{\#1}=\bm{P} \mathbbm{1}_{n_2}$ and $\bm{P}_{\#2}=\bm{P}^T \mathbbm{1}_{n_1}$ are the marginals of $\bm{P}$, 
  and $\text{KL}$ is the Kullback-Leibler divergence. The first two terms correspond to the FGW loss, 
  while the last term is a penalty that encourages the transport plan to be close to a balanced plan. 
  The parameter $\rho$ controls the strength of this penalty, with larger values of $\rho$ encouraging more 
  balanced plans. In practice, we sample $\rho$ from a log-uniform distribution on $[10^{-7}, 1]$ during 
  training to learn transport plans with different levels of unbalancedness, which adds variability to 
  the dataset and improves the generalization of the ULOT model.

\section{Experimental details}

Training for ULOT and dictionary learning are performed on one H100 GPU and all the hyperparameters are 
validated using the Optuna library \cite{optuna_2019}. We detail the hyperparameters used for both trainings 
in the following sections.
\subsection{Dictionary learning training}\label{sec:experimental_details}

Hyperparameters for dictionary learning 
are given in Table \ref{table:hyperparameters_DL}. 
Dictionary learning training takes about $10$ hours.

\begin{table}[h]
  \centering
  \caption{Dictionary learning hyperparameters}
  \begin{tabular}{lc}
  \hline
  \textbf{Hyperparameter}      &  \\
  \hline
  Learning rate outer loop              & $0.001$                  \\
  Learning rate inner loop               & $0.09$                     \\
  Batch size                   & $32$                       \\
  Optimizer                    & AdamW                        \\
  Weight decay outer loop            & $1e^{-4}$                      \\
  Weight decay inner loop            & $1e^{-7}$               \\

  \hline
  \end{tabular}
  \label{table:hyperparameters_DL}
  \end{table}

\begin{table}[h]
  \centering
  \caption{ULOT hyperparameters}
  \begin{tabular}{lc}
  \hline
  \textbf{Hyperparameter}      &   \\
  \hline
  Learning rate                & $0.002$                   \\
  Batch size                   & $64$                       \\
  Optimizer                    & AdamW                     \\
  Number of node embedding layers            & $3$                    \\
  Embedding dimension for  $\alpha$        & $16$                  \\
  Node embedding layer final out dimension & $64$ \\
  MLP hidden dimension & $1024$  \\
  GCN hidden dimension & $128$  \\
  Temperature value $a$ & $10$  \\
  \hline
  \end{tabular}
  \label{table:hyperparameters_ULOT}
\end{table}

\subsection{ULOT training}\label{sec:ULOT_convergence}

\begin{figure}[t]
  \centering
  \includegraphics[width=0.5\textwidth]{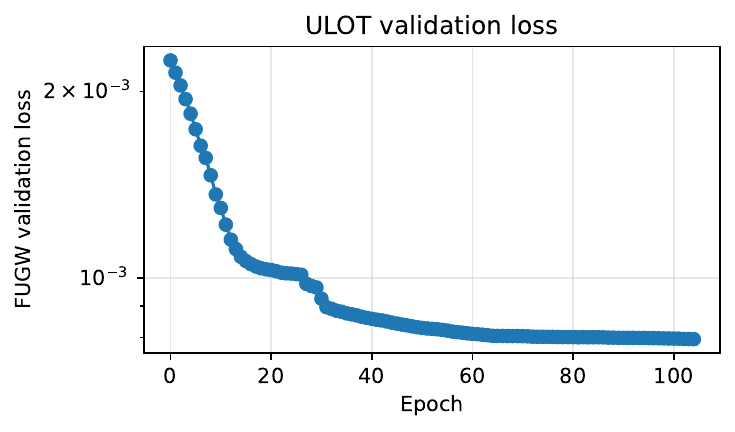}
  \caption{FUGW loss on validation dataset for ULOT training along epochs}
  \label{fig:ULOT_loss}
\end{figure}

ULOT is trained for $105$ epochs, and we show in Figure \ref{fig:ULOT_loss} the validation 
loss along epochs. Training takes approximately $200$ hours.
We use the hyperparameters in Table \ref{table:hyperparameters_ULOT}.

\subsection{Experimental details on contrasts}\label{sec:contrasts}

We derive contrast maps from signals in 
both native geometry and fsaverage space from the HCP dataset \cite{van2013wu}. 
For each task, we select the main contrasts reported in Table \ref{table:tasks} and 
restrict the analysis to the left hemisphere. Geodesic distances on the cortical surface 
are approximated by shortest-path distances on the corresponding graphs. The medial wall 
is excluded from all contrast maps.

\begin{table}[H]
  \centering
  \caption{HCP tasks used for alignment}
  \begin{tabular}{ll}
  \hline
  \textbf{Task} & \textbf{Contrast} \\
  \hline
  GAMBLING & win - loss \\
  EMOTION & fear - neutral \\
  MOTOR & left hand - average \\
  MOTOR & right hand - average \\
  MOTOR & left foot - average \\
  MOTOR & right foot - average\\
  RELATIONAL & match - relation \\
  WM & 2-back average - 0-back average \\
  SOCIAL & mental - random \\
  LANGUAGE & story - math \\
  \hline
  \end{tabular}
  \label{table:tasks}
\end{table}

\section{Additional visualizations}

\subsection{Dictionary learning atoms visualization}\label{sec:atoms}

\begin{figure}[H]
  \centering
  \includegraphics[width=0.85\textwidth]{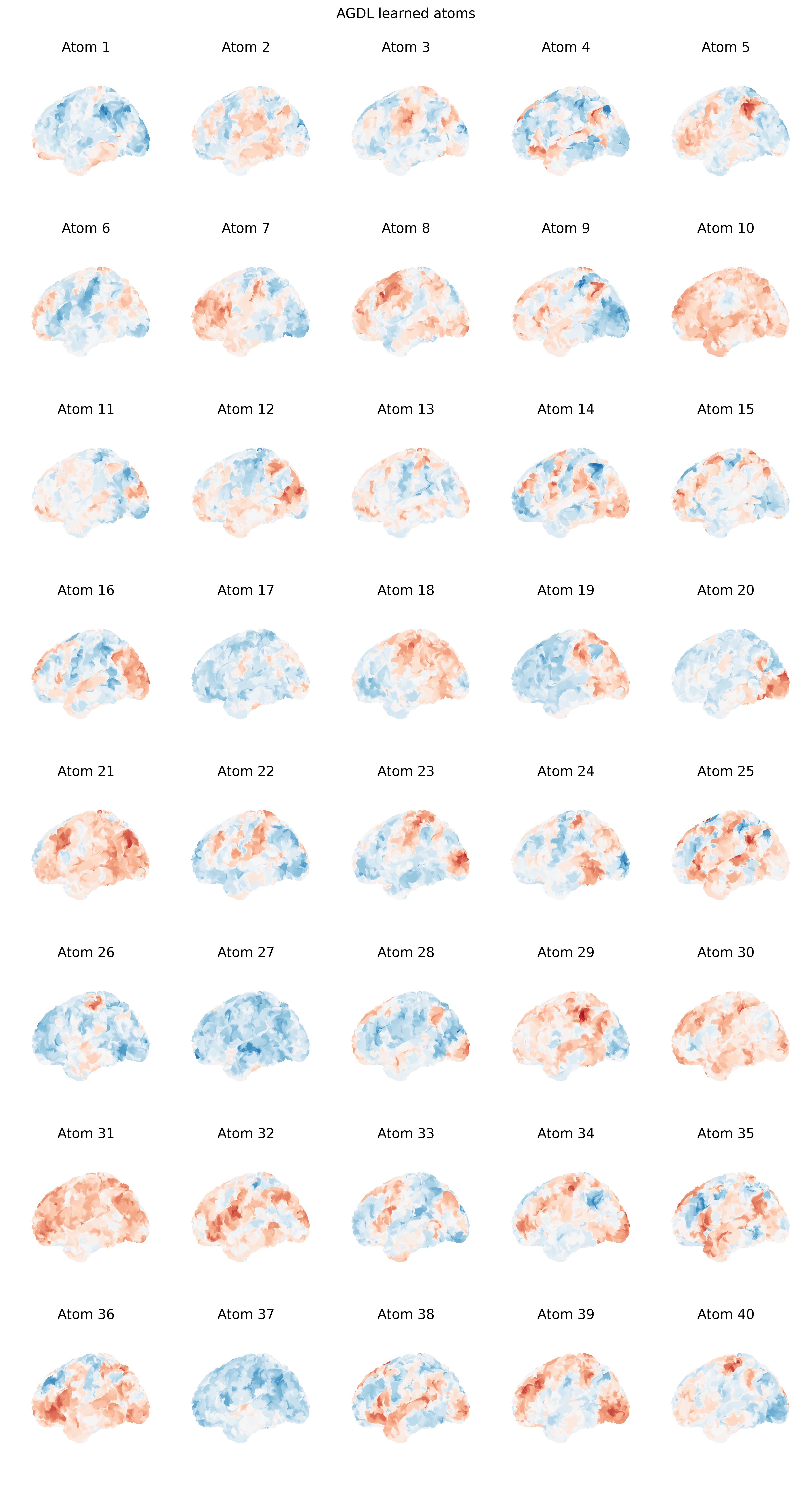}
  \caption{All learned dictionary atoms with AGDL for $\alpha=0.78$ (best subject classification performance)}
  \label{fig:al_atoms}
\end{figure}

We visualize all the atoms learned with AGDL for $\alpha=0.78$ (best subject classification performance) in Figure \ref{fig:al_atoms}. We see that many atoms have sharp activations in specific 
regions, while some others have more distributed activations.

\subsection{Additional visualizations of the learned representations}\label{sec:embeddings}

\begin{figure}[t]
  \centering
  \includegraphics[width=0.5\textwidth]{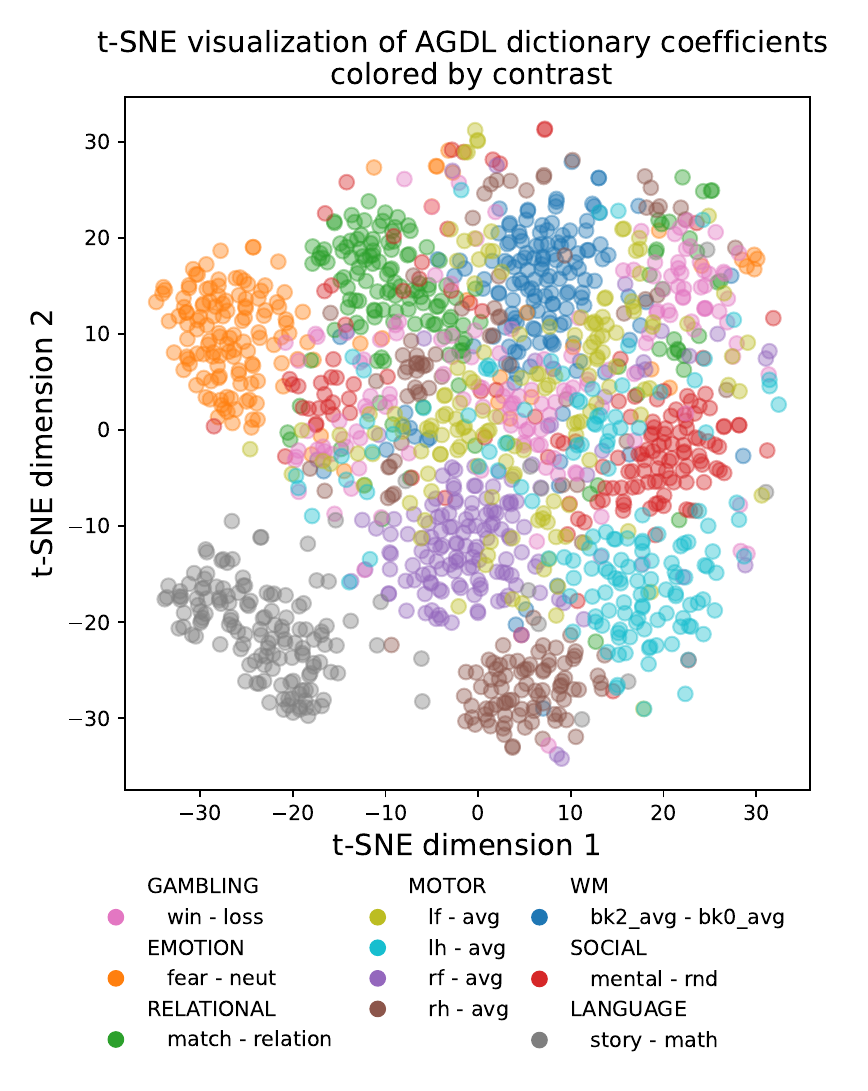}
  \caption{t-SNE visualization of the learned embeddings for all contrasts for $\alpha=0.22$.}
  \label{fig:tsne_all}
\end{figure}

We visualize the learned representations for all contrasts in Figure \ref{fig:tsne_all} 
with a t-SNE colored by the contrasts. Compared to Figure \ref{fig:tsne}, we add
contrasts \textit{left hand - average} and \textit{left foot - average} which are 
mostly activated in the right hemisphere and therefore are more difficult to discriminate.
We observe that the representations learn less meaningful representations on these contrasts on
Figure \ref{fig:tsne_all} compared to the other contrasts.

\begin{figure}[H]
  \centering
  \includegraphics[width=\textwidth]{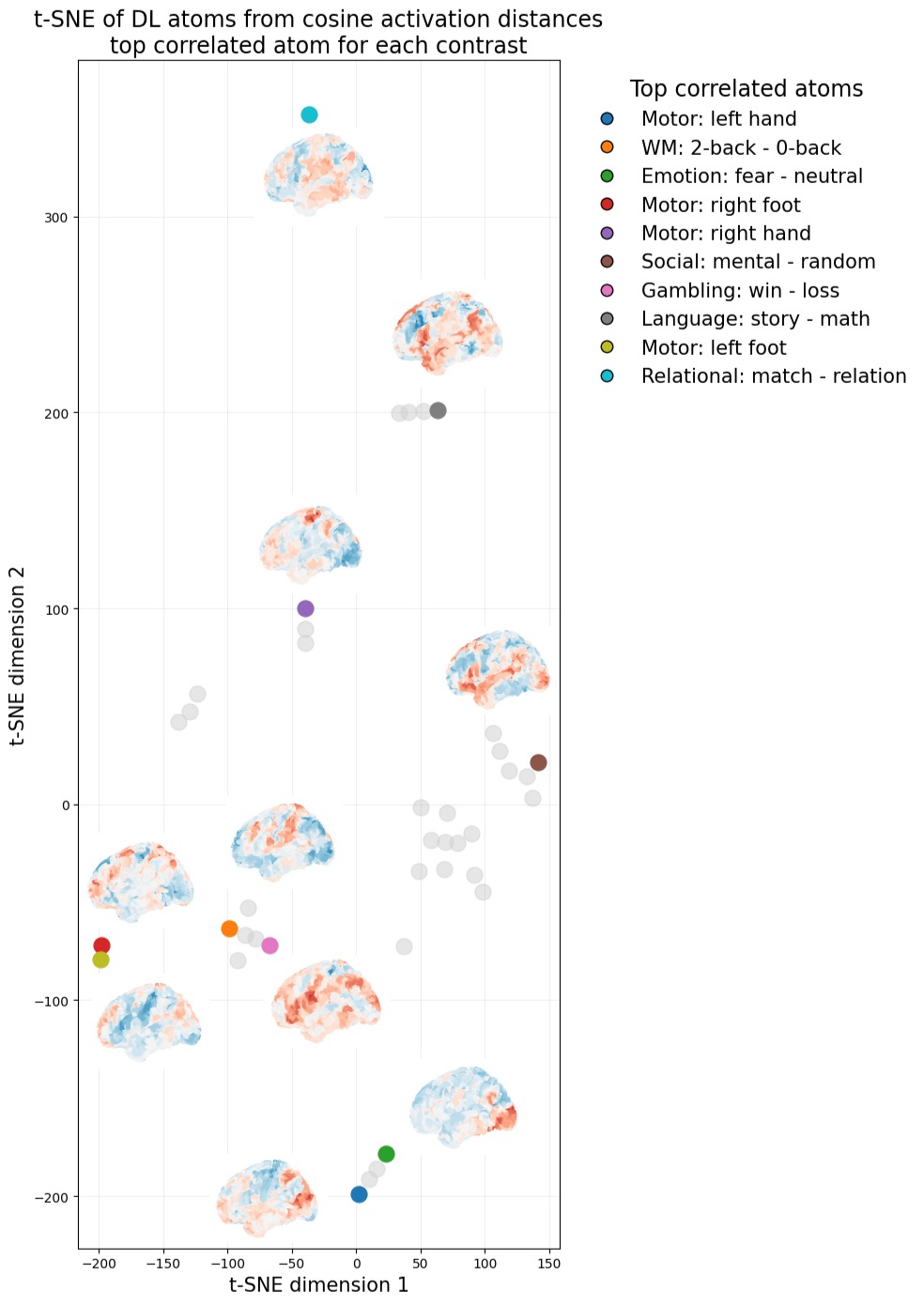}
  \caption{t-SNE visualization of the learned atoms, where each point corresponds 
  to an atom. The $10$ colored atoms are associated to the contrast they are most correlated with,
  and are plotted next to their visualization.}
  \label{fig:tsne_atoms}
\end{figure}

We also visualize the atoms by doing a t-SNE on the distance matrix between the atoms, 
where the distance between two atoms is defined as 
the cosine distance between their activations across the dataset. 
We show in Figure \ref{fig:tsne_atoms} the t-SNE visualization of
 the atoms for $\alpha=0.78$ (best subject classification performance), 
 where each point corresponds to an atom.
 The $10$ atoms most correlated with each contrast are colored.
 We observe that the $10$ most correlated atoms span the whole space of atoms
and that they have activations consistent with the contrast they are associated to.

\end{document}